\documentclass[conference,a4paper]{IEEEtran}

\IEEEoverridecommandlockouts
\usepackage{cite}
\usepackage{amsmath,amssymb,amsfonts}
\usepackage{algorithmic}
\usepackage{graphicx}
\usepackage{textcomp}
\usepackage{xcolor}
\usepackage{subfig}
\usepackage{url}

\newcommand{\etal}{\textit{et al}. }
\newcommand{\ie}{\textit{i}.\textit{e}. }
\newcommand{\eg}{\textit{e}.\textit{g}. }

\def\BibTeX{{\rm B\kern-.05em{\sc i\kern-.025em b}\kern-.08em
    T\kern-.1667em\lower.7ex\hbox{E}\kern-.125emX}}
\begin{document}

\title{Non-blind Image Restoration Based on Convolutional Neural Network}

\author{\IEEEauthorblockN{Kazutaka Uchida\IEEEauthorrefmark{1}, Masayuki Tanaka\IEEEauthorrefmark{1}\IEEEauthorrefmark{2}, and Masatoshi Okutomi\IEEEauthorrefmark{1}}
\IEEEauthorblockA{
\IEEEauthorrefmark{1}\textit{Tokyo Institute of Technology},
2--12--1 Ookayama, Meguro--ku, Tokyo, Japan \\
uchida@ok.sc.e.titech.ac.jp, \{mtanaka,mxo\}@sc.e.titech.ac.jp}
\IEEEauthorblockA{\IEEEauthorrefmark{2}\textit{National Institute of Advanced Industrial
 Science and Technology}, 
2--3--26 Aomi, Koto--ku, Tokyo, Japan
}}

\maketitle

\begin{abstract}
Blind image restoration processors based on convolutional neural network (CNN) are intensively researched because of their high performance. However, they are too sensitive to the perturbation of the degradation model. They easily fail to restore the image whose degradation model is slightly different from the trained degradation model. In this paper, we propose a non-blind CNN-based image restoration processor, aiming to be robust against a perturbation of the degradation model compared to the blind restoration processor. Experimental comparisons demonstrate that the proposed non-blind CNN-based image restoration processor can robustly restore images compared to existing blind CNN-based image restoration processors.



\end{abstract}


\section{Introduction}

A convolutional neural network (CNN) can handle a wide variety of image processing such as colorization \cite{Iizuka:2016:LCJ:2897824.2925974, varga2016fully}, super-resolution \cite{dong2016image, kim2016deeply, shi2016real,kim2016accurate}, noise reduction \cite{jain2009natural}, and JPEG deblocking \cite{yu2016deep, li2017efficient}.
Various types of network structures have been proposed \cite{he2016identity,ronneberger2015u,uchida2018coupled}.
One of the reasons that CNN image processors are paid enthusiastic attention is its simpleness of one-to-one direct mapping from the input image to the output image.
Just feeding ``before'' and ``after'' image dataset to a network for training, we can generate a practical image processor.

Kim \etal \cite{kim2016accurate} presented a CNN image processor called VDSR for single image super-resolution (SISR). VDSR is a plain feed-forward network which has 20 convolutional layers that utilizes residual learning. The training dataset consists of multi-scale low resolution images as ``before'' images and the corresponding high resolution images as ``after'' images. VDSR predicts the super-resolved image without explicit image prior, \ie scale factor, because it implicitly infers the scale factor of an input image.

Zhang \etal \cite{zhang2017beyond} extended VDSR to a model called DnCNN-3 that can handle three different degradation types, namely additive white Gaussian noise (AWGN), low-resolution, and JPEG compression block noise.
Structure of the model is similar to VDSR, though, the main improvement is the fed image dataset. It comprises ``before'' and ``after'' images of the three types of degradation. As a result, DnCNN-3 obtains ability to predict a cleaned image from three different types of degraded images without image prior on degradation.


Those image restoration processors are classified into a blind image restoration processor. The blind image restoration processor restores images with implicitly or explicitly estimating the degradation model and parameters.
Although learning-based blind restoration processors have very high performance, they are very sensitive to the perturbation of the degradation model. For instance, the learning-based blind-restoration processors easily fail to restore the image whose degradation model is slightly different from the trained degradation model.

In this paper, we propose a non-blind CNN-based image restoration processor.
The non-blind image restoration processor restores images with the degradation model and parameters given by users.
If we can train a blind restoration processor with all kind of degradation model, such a blind restoration processor is better than a non-blind restoration processor.
However, it is infeasible to train with all kind of degradation model. In contrast, we can say that a non-blind image restoration processor is robust against a perturbation of the degradation model compared to the blind restoration processor.


First, we describe image restoration from degradation with perturbation in Section \ref{sec:denoiser}. Then, we propose the non-blind CNN-based image restoration processor in Section \ref{sec:proposed}, and demonstrate experimental results in Section \ref{sec:experiments}.

\section{Limitation of Learning Based\\Image Restoration}

\label{sec:denoiser}

\subsection{Restoration from degradation with perturbation}

The CNN-based blind image restoration processors such as VDSR and DnCNN-3 do not require degradation information. However, those blind CNN-based image restoration processors only work, if the training dataset includes same statistical properties of input degradation image. In practice, actual image degradation has often different characteristics from that of the training dataset.

For example, in image capturing process, there are several degrading processes such as blurring, random noise with an image sensor, and JPEG compression.
If a denoiser is trained with random noise and a degraded image is randomly noised and JPEG compressed, the denoiser fails to restore the image because the degradation has different property from trained datasets. In this case, degradation by JPEG compression can be considered as perturbation against the trained degradation model (\ie random noise).






\subsection{Blind Restoration for non-trained degradation model}

\begin{figure}[!t]
  \captionsetup{farskip=0pt}
    \begin{center}
  \subfloat[Degraded]{%
       \includegraphics[width=0.24\linewidth]{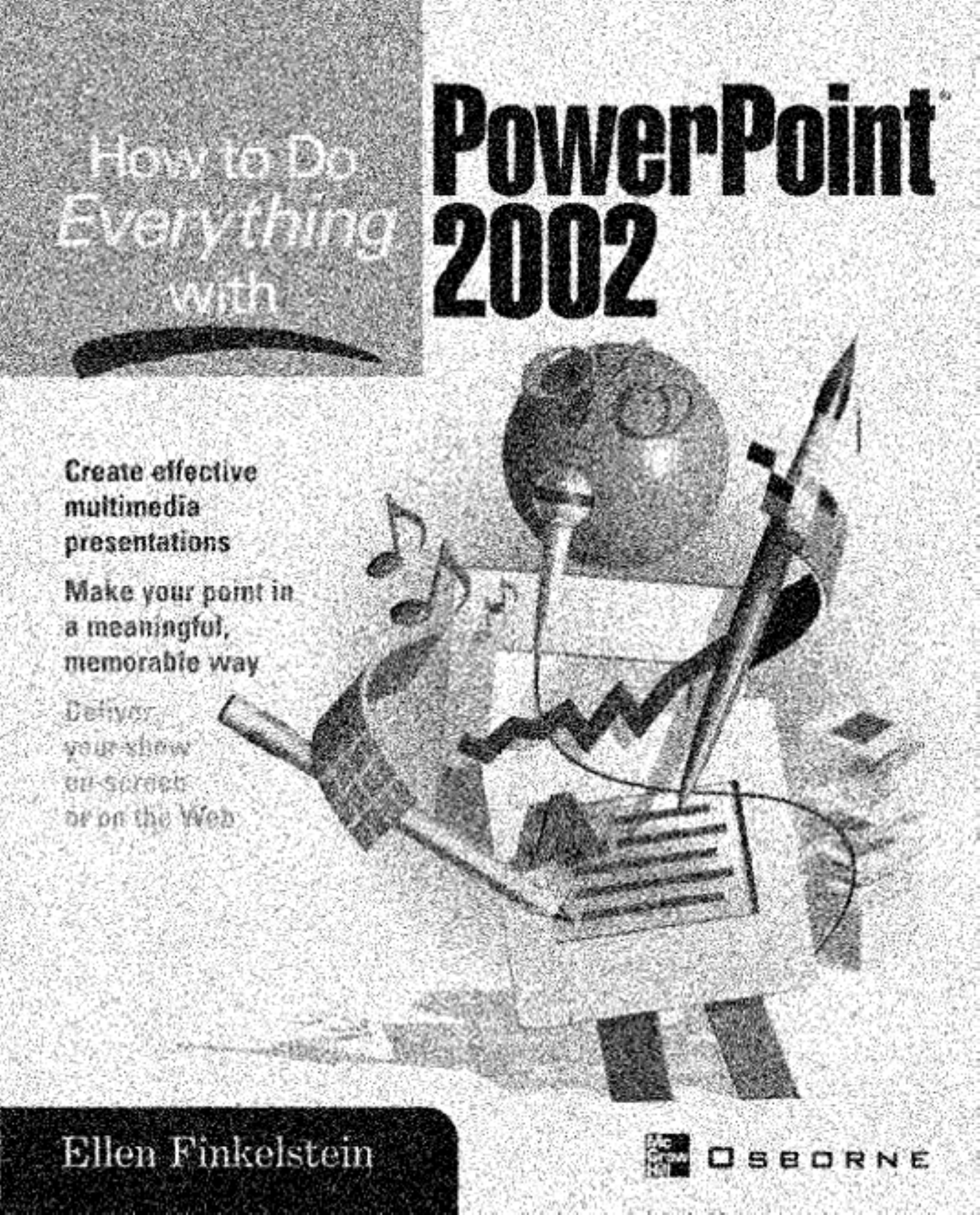}}
    \label{aua}\hfill
  \subfloat[DnCNN-3]{%
       \includegraphics[width=0.24\linewidth]{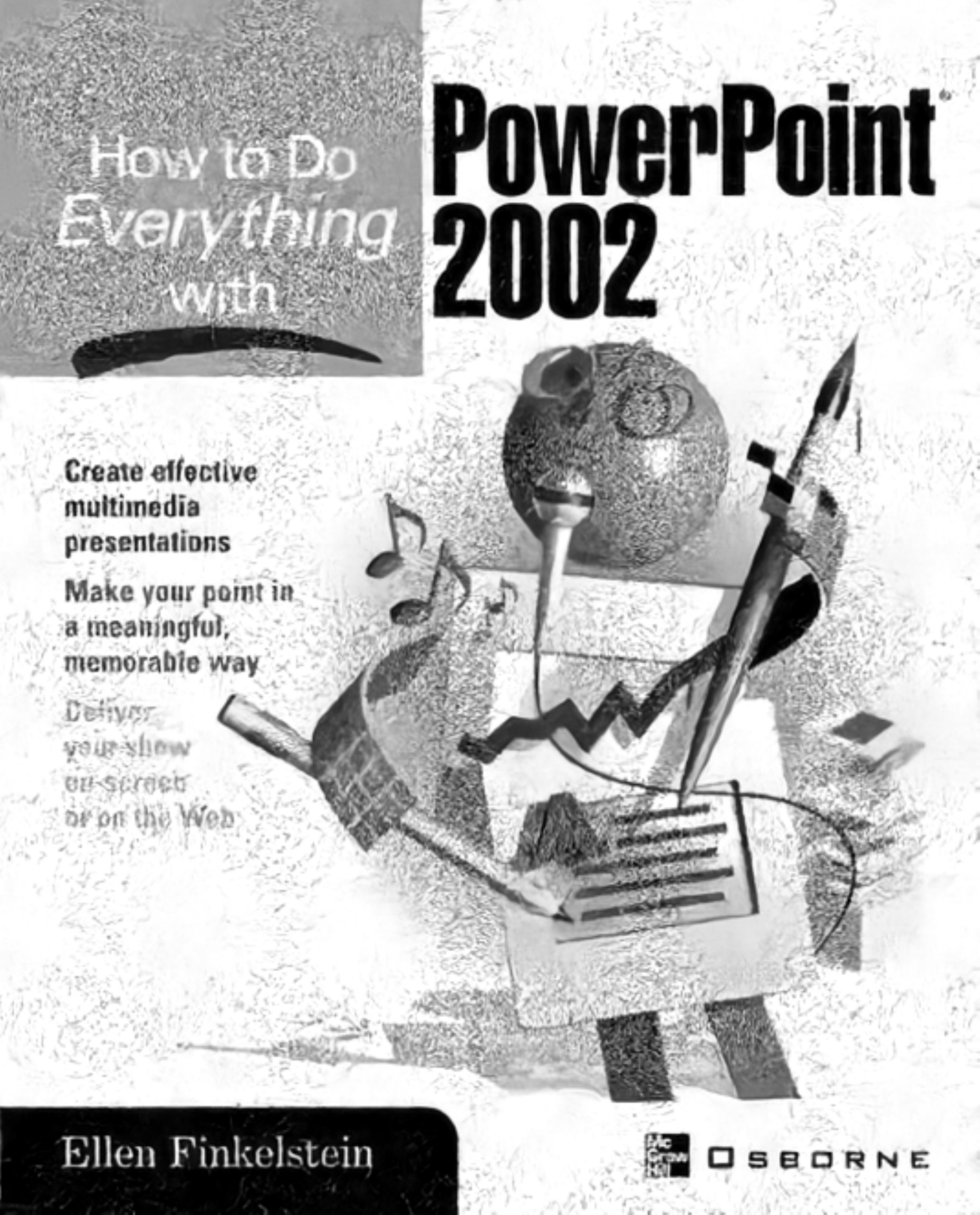}}
    \label{aub}\hfill
    \subfloat[BM3D]{%
         \includegraphics[width=0.24\linewidth]{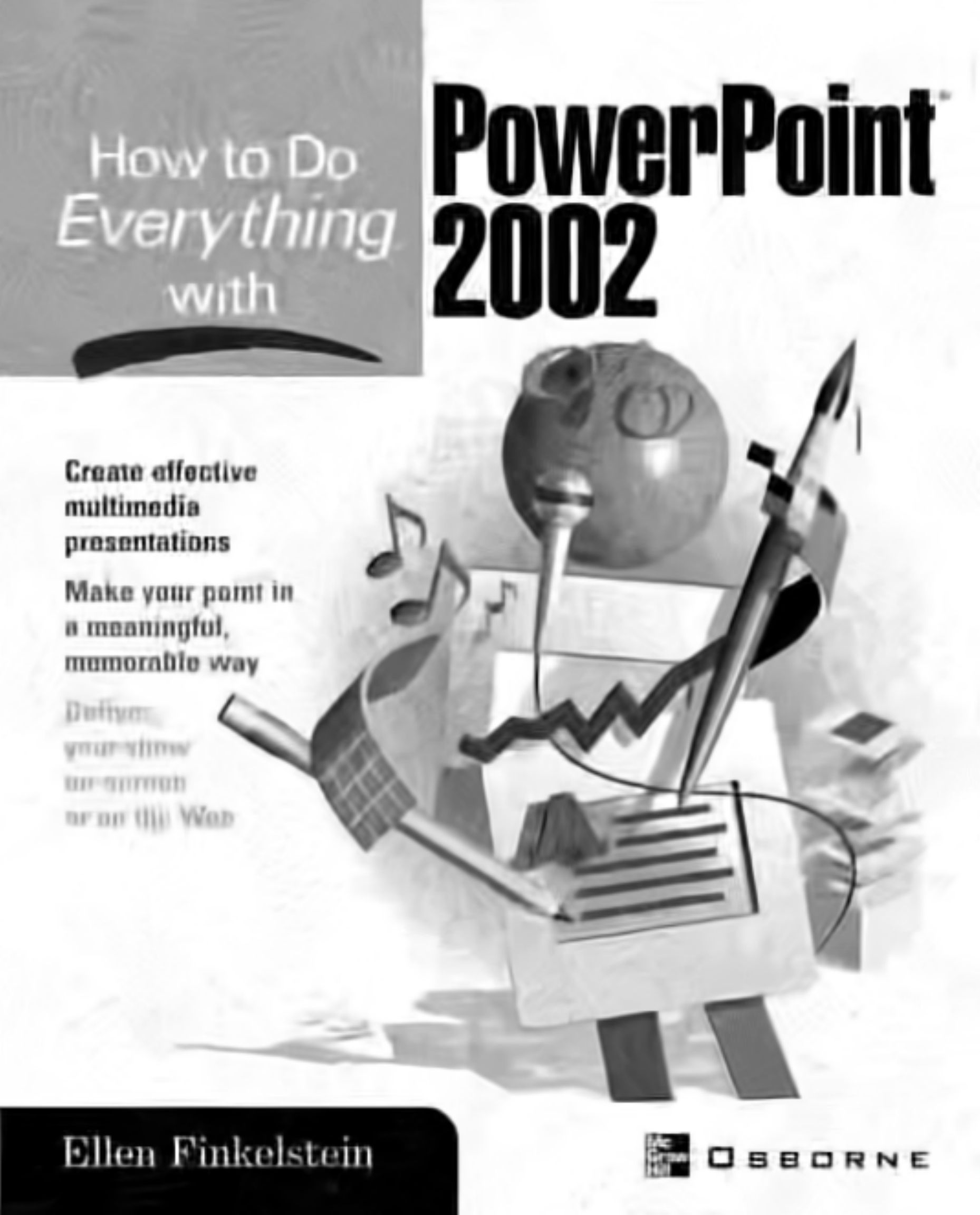}}
      \label{auc}\hfill
    \subfloat[Proposed]{%
         \includegraphics[width=0.24\linewidth]{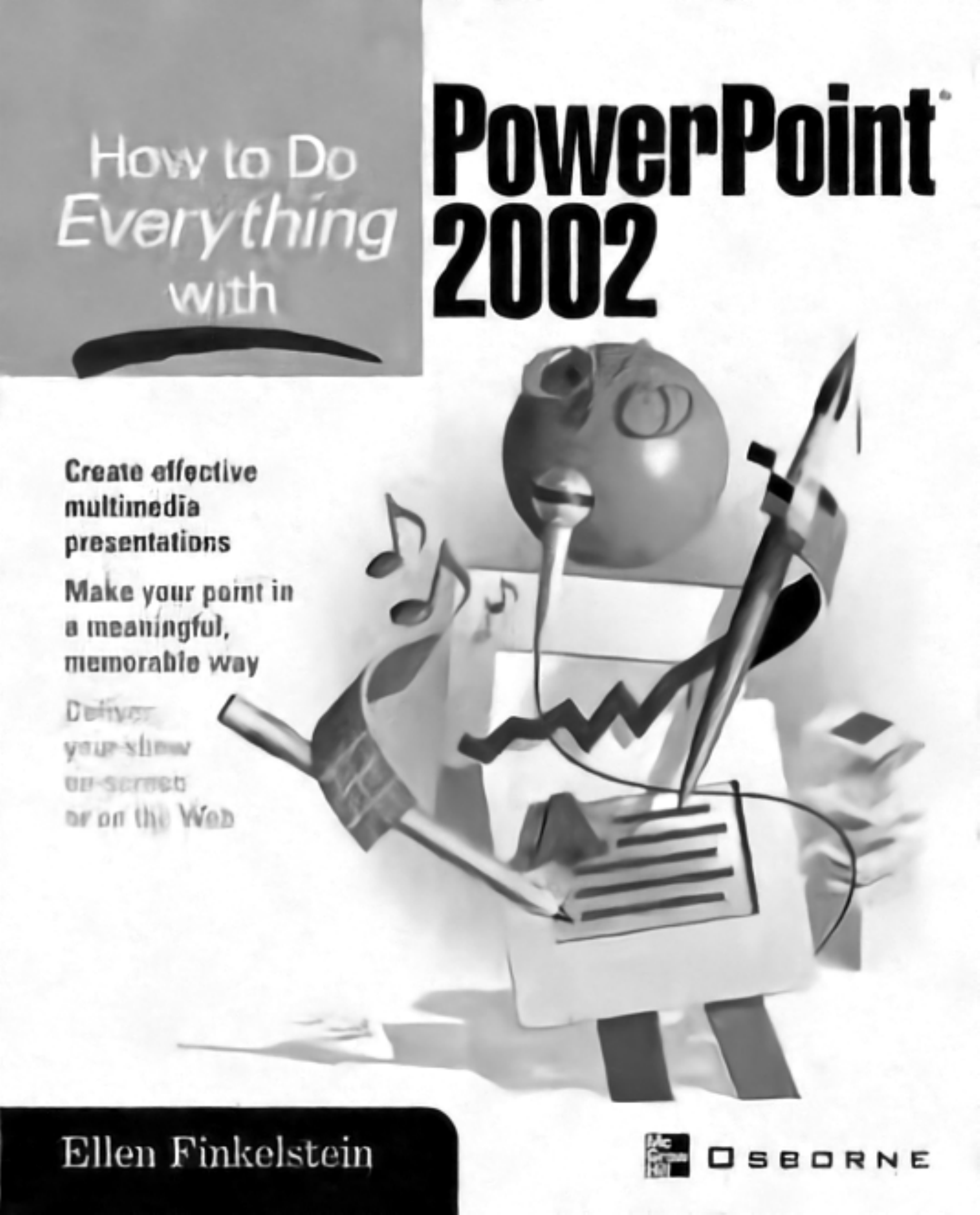}}
      \label{aud}\hfill
  \end{center}
  \caption{Restoration from AWGN + upscaling degradation}
  \label{fig:sample_result}
\end{figure}

Figure \ref{fig:sample_result} shows an example of restoration by DnCNN-3\footnote{Please note that DnCNN-3 model itself is expected to achieve higher performance if trained directly with the AWGN + upscaling degradation. This comparison is intended to demonstrate that CNN-based model performs poorly on images with untrained degradation model.}, BM3D, and our method.
The input image (Fig. \ref{fig:sample_result} (a)) are degraded with AWGN with $\sigma=50/255$ and then upscaled by 1\% in size.
The restored image by DnCNN-3 (Fig. \ref{fig:sample_result} (b)) has still AWGN in some regions.
This indicates that DnCNN-3 did not recognize the input image as AWGN image in some area because the processor does not know the degradation pattern.
As a result, the processor could not perform denoising, in some area.

We can consider that the degradation information is implicitly estimated in the blind CNN-based image restoration processor like DnCNN-3. If the degradation information is wrongly estimated, the blind CNN-based image restoration processor fails to restore the image.

It is also confirmed that model-based noise level estimation algorithms \cite{liu2012noise,liu2013single,liu2013estimation} fail to estimate true noise level when the input image has such perturbation.

\subsection{Controlling model by parameter}

To control the restoration strategy, it is useful to have input channels to give parameters as a guidance.
Chen \etal \cite{chen2017fast} proposed a fully convolutional network that approximates ten image operators such as ``style transfer'' \cite{Aubry14b}, ``Pencil drawing'' \cite{Lu:2012:CST:2330147.2330161}, and ``Nonlocal dehazing'' \cite{NonLocalImageDehazing}.
Their proposed model has parameter channels along with an input image to adjust properties for the image processing.
While the training phase, parameters used for generating ``after'' images from ``before'' images are set to the parameter channels.
We consider that the similar approach is applicable to image restoration application taking degradation attributes as the input to control the restoration strategy.

\section{Proposed Non-Blind Image Restoration}
\label{sec:proposed}

We propose a non-blind CNN-based image restoration processor.
Here, we describe network structure, training processor, and inference process of the proposed non-blind image restoration.

\subsection{Network Structure}

\begin{figure*}
\begin{center}
   \includegraphics[width=0.72\linewidth]{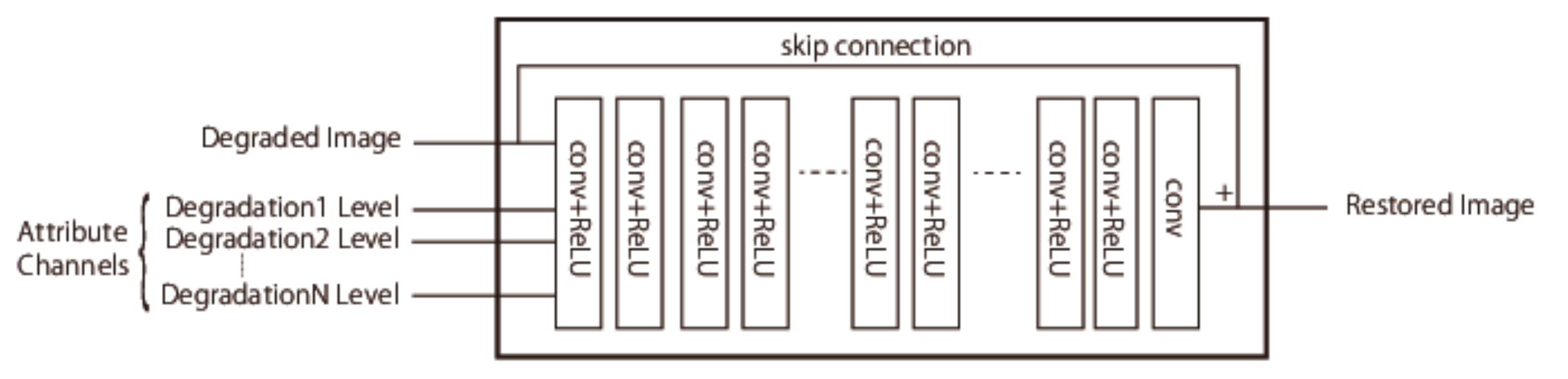}
\end{center}
   \caption{Proposed network structure: The network consists of $L$ blocks of a convolution layer and ReLU as the activation function. Inputs are a degraded image and $N$ degradation attribute channels. Restored image is output by adding the predicted residual and input image.}
\label{fig:network}
\end{figure*}

The network structure of the proposed non-blind CNN-based image restoration processor is shown in Fig. \ref{fig:network}. The proposed network includes the skip connection as same as in the VDSR network \cite{kim2016accurate}. Following \cite{chen2017fast}, we added $N$ attribute channels to the degradation image as input data. The width and height of the attribute channels are same as those of the degradation image.

The network model consists of $L$ convolutional layer, typically $L=20$, with $3 \times 3$-sized filter with 64 feature maps. We use a ReLU (Rectified Linear Unit) for an activation function. For the last layer, we use no activation function to output residual between the reconstructed image and the input image.

Intensity range of input and output image is normalized to $[0,1]$. Parameters of the degradation attribute are also normalized to $[0,1]$ which represent degradation magnitude for each degradation model.




\subsection{Training Process}

Training dataset is generated from clean images by degrading each image with a random degradation parameter for one of the $N$ degradation types.
The mean squared-error loss function is used, where the error is the average of pixel-wise squared differences between the predicted image and the ground truth.
The optimization of the model can be done using general optimization method such as stochastic gradient descent (SGD) or Adam \cite{kingma2014adam}.

\subsection{Inference Process}

Inference is a simple feed-forward process.
The non-blind restoration requires information on degradation attribute of the degraded image in the degradation attribute channels. The model is expected to use the information to decide the restoration strategy, because the information was always true with samples while training. Hence the degradation attribute channels are used as degradation parameters for the non-blind restoration.

\section{Experiments}
\label{sec:experiments}


\subsection{Model Settings}

The number of layers $L$ is set to 20 for the proposed network.
For degradation types, AWGN, low-resolution, and JPEG compression are assumed.
Thus, the input channels have one input image channel and three attribute channels.


Input images are Gaussian noised, low-resolution, or JPEG compressed image. The Gaussian noise level is ranging from $\sigma=5/255$ to $55/255$, the scale factors of resolution enhancement or super-resolution are $\times 1, \times 2, \times 3, \times 4$, and JPEG quality of compressed image is between 5\% to 95\%.

\begin{figure}[t!]
  \captionsetup{farskip=0pt}
    \begin{center}
  \subfloat[Gaussian noise]{%
       \includegraphics[width=0.33\linewidth]{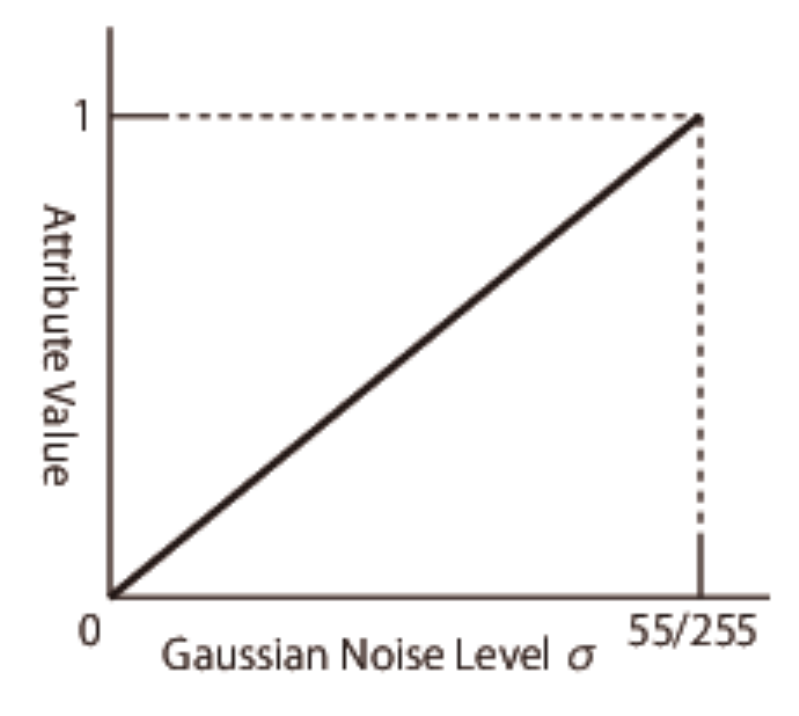}}
    \label{5a}\hfill
  \subfloat[Scale Factor]{%
       \includegraphics[width=0.33\linewidth]{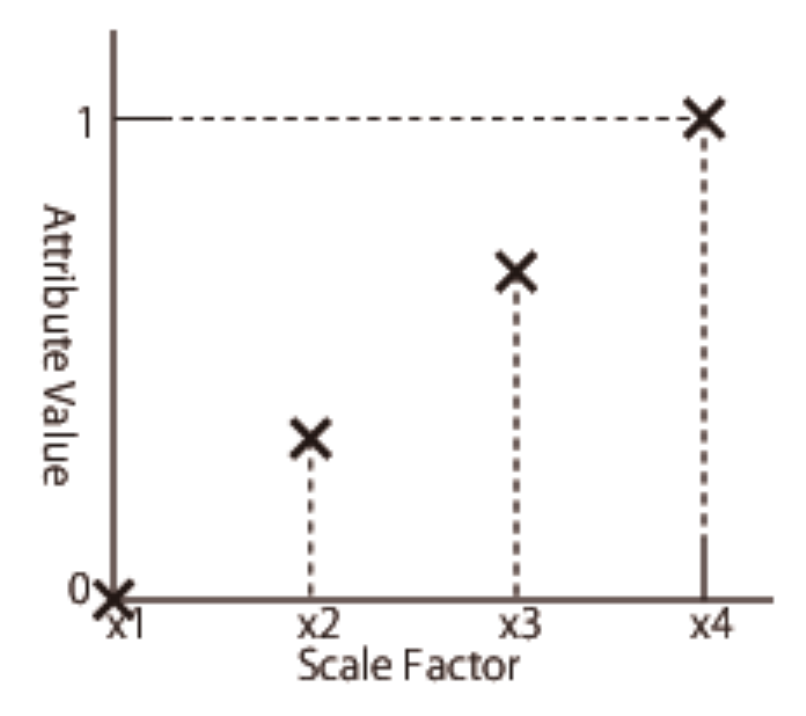}}
    \label{5b}\hfill
  \subfloat[JPEG Quality]{%
       \includegraphics[width=0.33\linewidth]{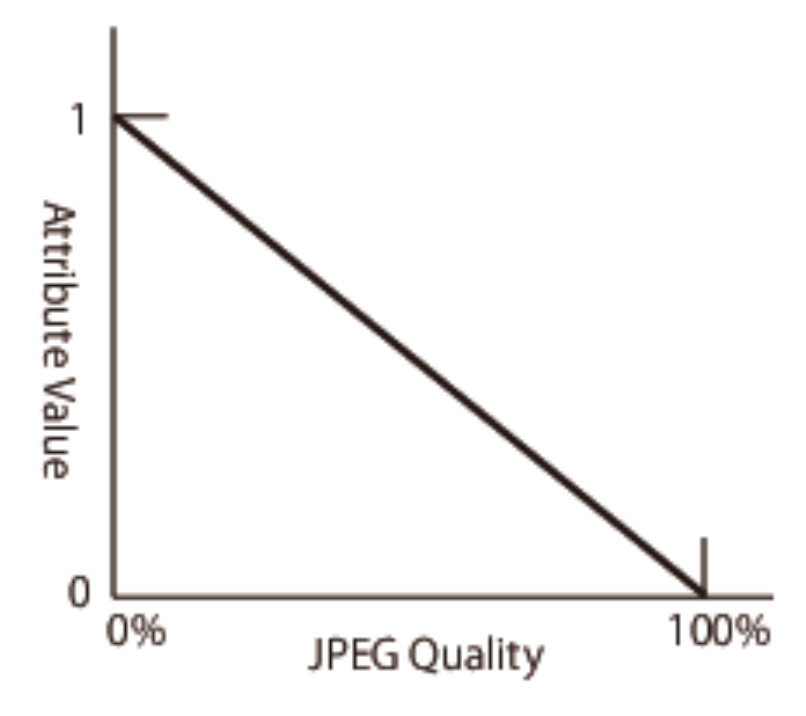}}
    \label{5c}
  \end{center}
  \caption{Values for the attribute channel.}
  \label{fig:attr_channels}
\end{figure}

Values for attribute channels are set as shown in Fig. \ref{fig:attr_channels}.
The first attribute channel represents Gaussian noise magnitude of the input image. The value of the channel are linearly proportional to the Gaussian noise magnitude $\sigma$ of the input image, that is zero when $\sigma=0/255$ and one when $\sigma=55/255$. The second attribute channel contains information on scale factor of the input image. The value of the channel is set such that zero for scale factor $\times 1$ and one for $\times 4$. The third attribute channel retains block noise magnitude of the input image. The value of the channel is proportional to the inverse of JPEG quality, that is zero for JPEG quality of 100\% and one for 0\%.

\subsection{Training}

Training dataset is generated from 291 images from Yang \etal \cite{yang2010image} and Berkeley Segmentation Dataset \cite{martin2001database}. The images are cropped to patches of size $50 \times 50$ with random degradation.
Values for attribute channels are set to the true degrading parameters for each sample.

The training samples are Gaussian noised, low-resolution, and JPEG compressed images. Parameters for noise level, scale factor, and JPEG quality are randomly chosen for each sample and there is no image sample compositing of different degradation types, \eg there is no image having both Gaussian noise and JPEG block noise.





The initial weights of the model are imported from that of DnCNN-3. Since the number of input channel are increased from DnCNN-3, small random weight are set for additional filters at the first layer.
The optimization is done by Adam in two steps.
Firstly, only layer 1 to 5 are trained for 10 epochs, where one epoch has 1024k data samples. Then, all the layers are trained for 80 epochs.
The implementation code is written with Keras framework \cite{chollet2015keras} and run on NVIDIA TITAN X GPU.

\subsection{Image restoration with trained degradation model}

As a preliminary examination, restoration performance of the trained model for single degradation is measured.
The comparison is demonstrated in Table \ref{tab:psnr}.
The same testing datasets are used as in \cite{zhang2017beyond}.

As for Gaussian noise denoising, the proposed model shows better accuracy because the model exploits given attribute on noise level for denoising.
This tendency is also true for JPEG deblocking.

The performance on super-resolution is constant between DnCNN-3 (blind) and our non-blind restorations.
This is because the scale factors are discrete and DnCNN-3 model can accurately estimate the scale factor from an input image.

\begin{table}
\begin{center}

\caption{Comparison of PSNR for the three degradation types}
\label{tab:psnr}

{

\begin{tabular}{|c|c|c|c|c|} \hline
\multicolumn{5}{|c|}{\bf{Gaussian Denoising}}\\ \hline
        & Noise & DnCNN-3 & BM3D & Ours (non-blind) \\ \cline{3-5}
Dataset & Level & PSNR / SSIM & PSNR / SSIM & PSNR / SSIM \\ \hline
        & 15/255 & 31.46 / 0.8826 & 31.07 / 0.8717 & 31.57 / 0.8874 \\
BSD68   & 25/255 & 29.02 / 0.8190 & 28.57 / 0.8013 & 29.11 / 0.8236 \\
        & 50/255 & 26.10 / 0.7076 & 25.62 / 0.6864 & 26.16 / 0.7129 \\ \hline
\end{tabular}

\vspace*{3 mm}

\begin{tabular}{|c|c|c|c|} \hline
\multicolumn{4}{|c|}{\bf{Single Image Super Resolution}}\\ \hline
        & Scale & DnCNN-3 & Proposed \\ \cline{3-4}
Dataset & Factor & PSNR / SSIM & PSNR / SSIM \\ \hline
        & 2 & 33.03 / 0.9128 & 32.97 / 0.9125 \\
Set14   & 3 & 29.81 / 0.8321 & 29.80 / 0.8323 \\
        & 4 & 28.04 / 0.7672 & 27.99 / 0.7660 \\ \hline
        & 2 & 30.74 / 0.9139 & 30.72 / 0.9141 \\
Urban100& 3 & 27.15 / 0.8276 & 27.12 / 0.8278 \\
        & 4 & 25.20 / 0.7521 & 25.17 / 0.7521 \\ \hline
\end{tabular}

\vspace*{3 mm}

\begin{tabular}{|c|c|c|c|} \hline
\multicolumn{4}{|c|}{\bf{JPEG Image Deblocking}}\\ \hline
        & Quality & DnCNN-3 & Proposed \\ \cline{3-4}
Dataset & Factor & PSNR / SSIM & PSNR / SSIM \\ \hline
        & 10 & 29.40 / 0.8026 & 29.41 / 0.8022 \\
Classic5& 20 & 31.63 / 0.8610 & 31.63 / 0.8610 \\
        & 30 & 32.91 / 0.8861 & 32.91 / 0.8861 \\
        & 40 & 33.77 / 0.9003 & 33.77 / 0.9004 \\ \hline
        & 10 & 29.19 / 0.8123 & 29.22 / 0.8120 \\
LIVE1   & 20 & 31.59 / 0.8802 & 31.60 / 0.8804 \\
        & 30 & 32.98 / 0.9090 & 32.99 / 0.9091 \\
        & 40 & 33.96 / 0.9247 & 33.97 / 0.9249 \\ \hline
\end{tabular}

}

\end{center}
\end{table}

\subsection{Robustness against perturbation of degradation model}

In order to evaluate the robustness against degradation model perturbation, we consider four different degradation processes: JPEG compression after adding AWGN, upscaling after adding AWGN, adding salt-and-pepper noise, and upscaling after JPEG compression.

\subsubsection{JPEG compression after adding AWGN}

\begin{figure}[!t]
  \captionsetup{farskip=0pt}
    \begin{center}
  \subfloat[Degraded]{%
       \includegraphics[width=0.24\linewidth]{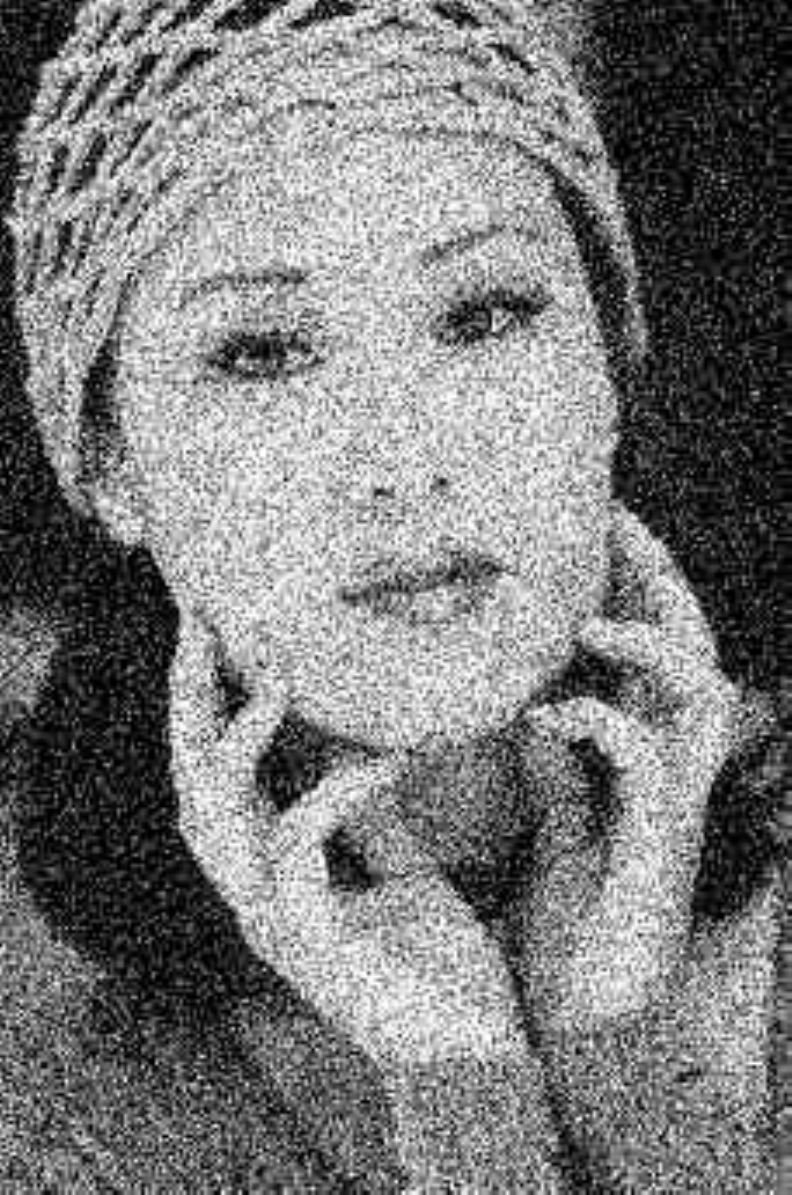}}
    \label{10b}\hfill
  \subfloat[DnCNN-3]{%
       \includegraphics[width=0.24\linewidth]{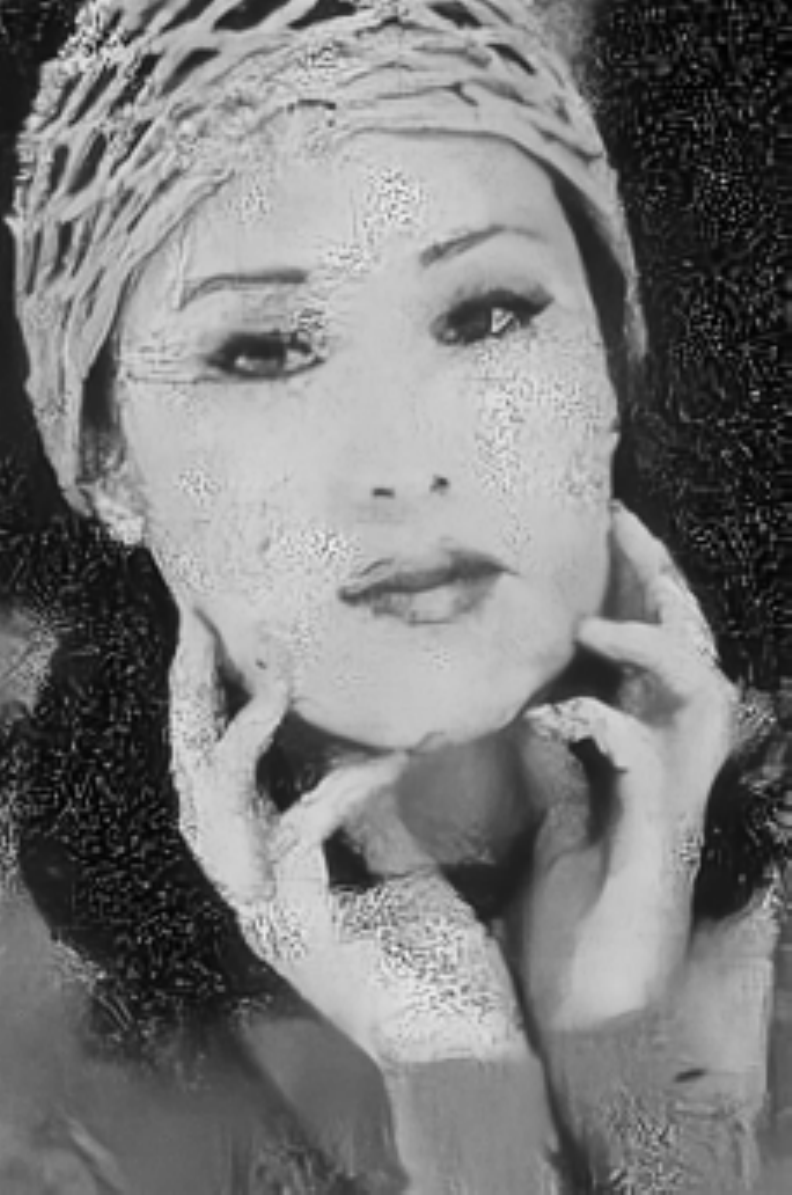}}
    \label{10c}\hfill
    \subfloat[BM3D]{%
         \includegraphics[width=0.24\linewidth]{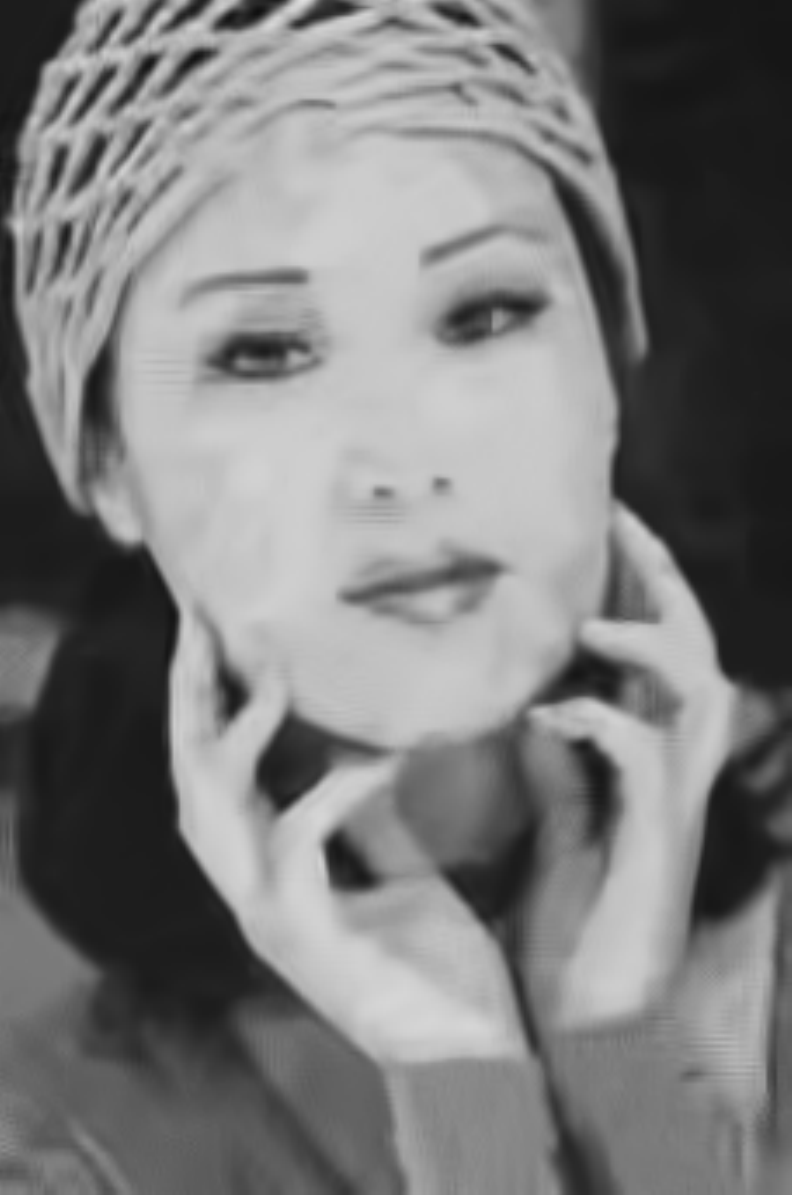}}
      \label{10e}\hfill
    \subfloat[Proposed]{%
         \includegraphics[width=0.24\linewidth]{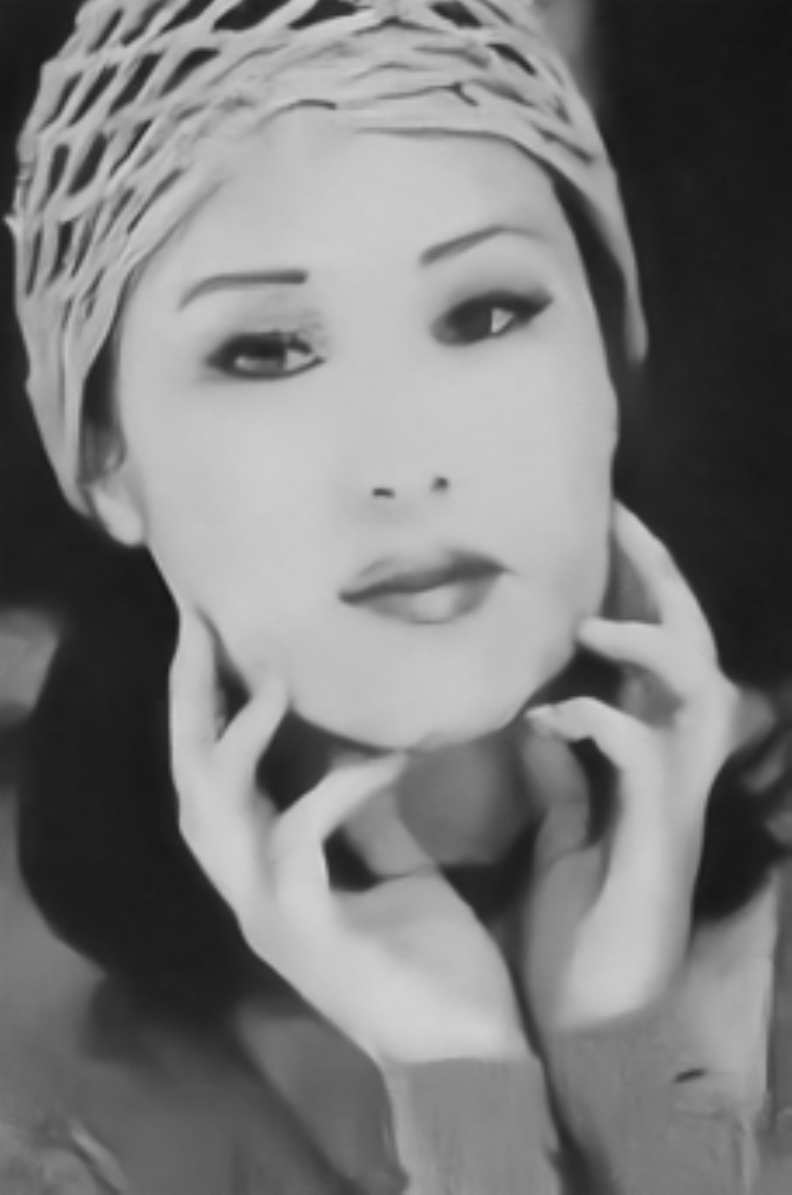}}
      \label{10e}\hfill
  \end{center}
  \caption{Restoration from AWGN + JPEG degradation}
  \label{fig:composite_result}
\end{figure}

The images are firstly degraded by adding AWGN whose noise level is $50/255$. Then, the JPEG compression with quality=30\% is applied. The restoration performances are shown in Table \ref{tab:psnr_untrained_awgn_jpeg}. The proposed non-blind CNN-based image restoration successfully achieves high PSNR compared to DnCNN-3 (blind/leaning-based) and BM3D (non-blind/non-learning-based). Figure \ref{fig:composite_result} shows examples of the restoration results.

\begin{table}[t!]
\begin{center}

\caption{Restoration results for AWGN + JPEG degradation}

\label{tab:psnr_untrained_awgn_jpeg}

\begin{tabular}{|c|c|c|c|} \hline
 & DnCNN-3 & BM3D & Proposed \\ \hline
Dataset & PSNR / SSIM & PSNR / SSIM & PSNR / SSIM \\ \hline
Set5 & 23.14 / 0.5875 & 25.17 / 0.7336 & {\bf 25.48 / 0.7387} \\ \hline
Set14 & 23.48 / 0.5997 & 24.74 / 0.6803 & {\bf 24.86 / 0.6870} \\ \hline
LIVE1 & 23.82 / 0.6077 & 24.67 / 0.6780 & {\bf 24.99 / 0.6942} \\ \hline
\end{tabular}

\end{center}
\end{table}

\subsubsection{Upscaling after adding AWGN}

The images are firstly degraded by adding AWGN whose noise level is $50/255$. Then, the upscaling by 1\% is applied. The restoration performances are shown in Table \ref{tab:psnr_untrained_awgn_upscaling}. As same as the case of the JPEG compression after adding AWGN, the proposed non-blind CNN-based image restoration successfully reduces the noise.




\begin{table}[t!]
\begin{center}

\caption{Restoration results for AWGN + upscaling degradation}

\label{tab:psnr_untrained_awgn_upscaling}

\begin{tabular}{|c|c|c|c|} \hline
 & DnCNN-3 & BM3D & Proposed \\ \hline
Dataset & PSNR / SSIM & PSNR / SSIM & PSNR / SSIM \\ \hline
Set5 & 21.26 / 0.4608 & 27.86 / 0.7944 & {\bf 28.42 / 0.8125} \\ \hline
Set14 & 20.86 / 0.4468 &  26.83 / 0.7362 & {\bf 27.08 / 0.7388} \\ \hline
LIVE1 & 20.64 / 0.4236 &  26.26 / 0.7188 & {\bf 26.66 / 0.7299} \\ \hline
\end{tabular}

\end{center}
\end{table}

\subsubsection{Salt-and-pepper noise}

The properties of the salt-and-pepper noise are very different from those of the AWGN. As shown in Table \ref{tab:psnr_snp}, the DnCNN-3 fails to reduce the noise, because the DnCNN-3 did not learn with the salt-and-pepper noise. The proposed non-blind image restoration processor achieved clearly higher PSNR than that of DnCNN-3 even though the proposed non-blind image restoration processor did not learn with salt-and-pepper noise.

BM3D exhibits higher performance than the proposed model and DnCNN-3 because it is originally designed for such noise patterns \cite{djurovic2016bm3d}.




\begin{figure}[t]
  \captionsetup{farskip=0pt}
    \begin{center}
  \subfloat[Input (salt-and-pepper noise)]{%
       \includegraphics[width=0.36\linewidth]{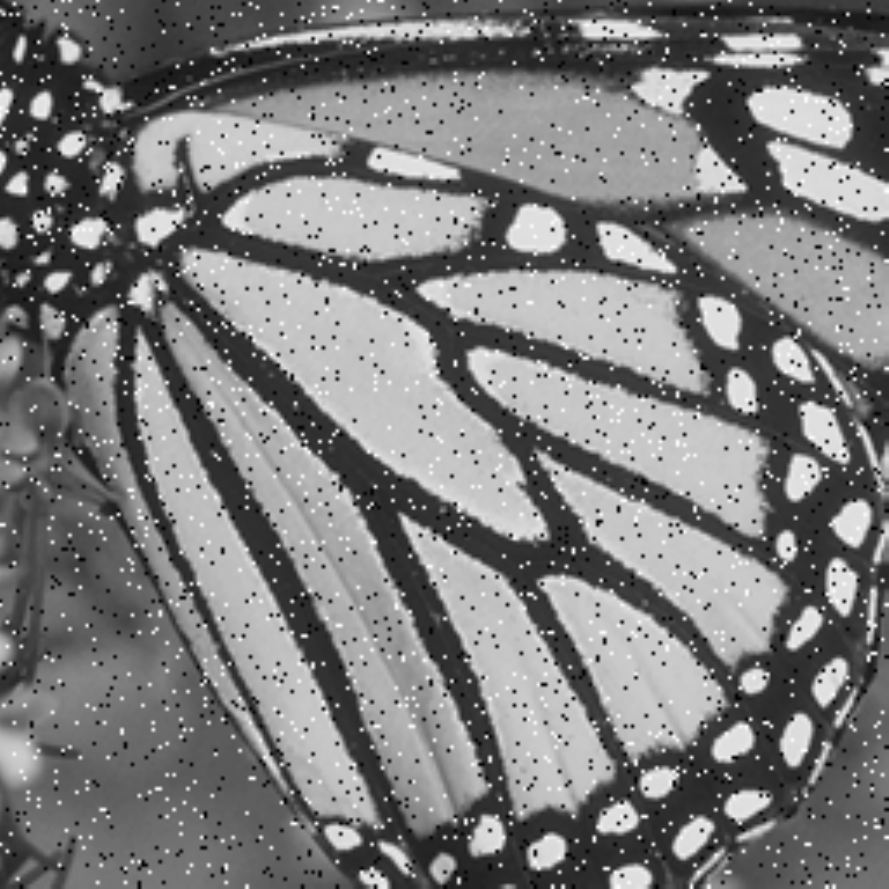}}
    \label{7a}\hspace{3mm}%
  \subfloat[DnCNN-3 (blind)]{%
       \includegraphics[width=0.36\linewidth]{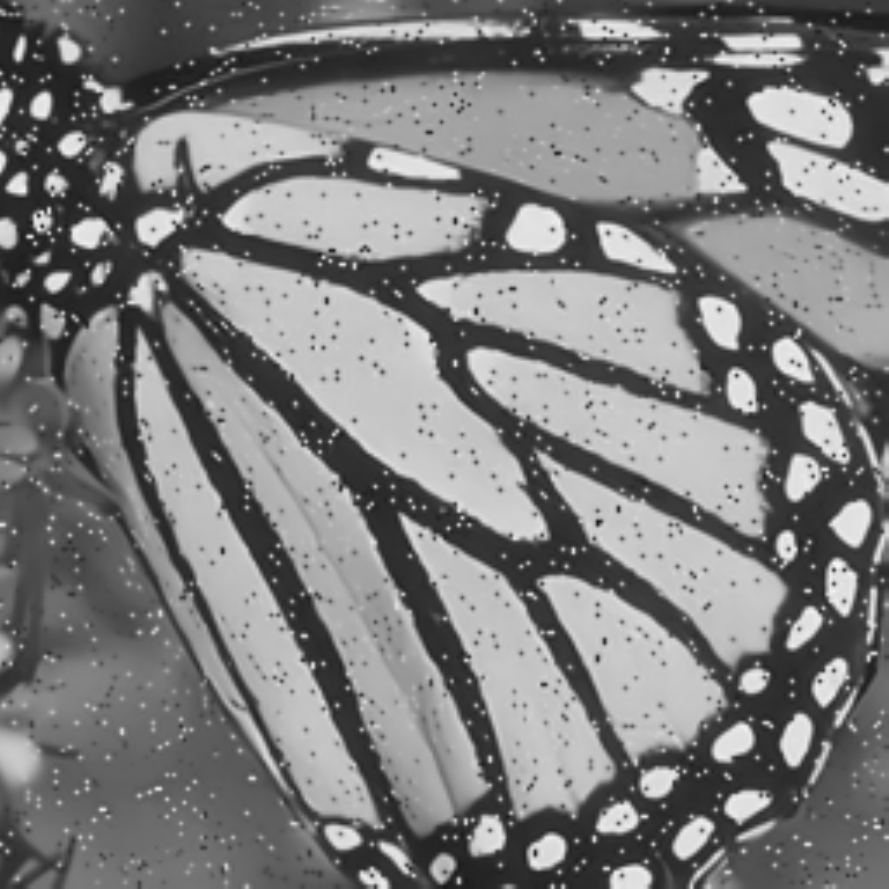}}
    \label{7b}
    \\
  \subfloat[BM3D (non-blind)]{%
         \includegraphics[width=0.36\linewidth]{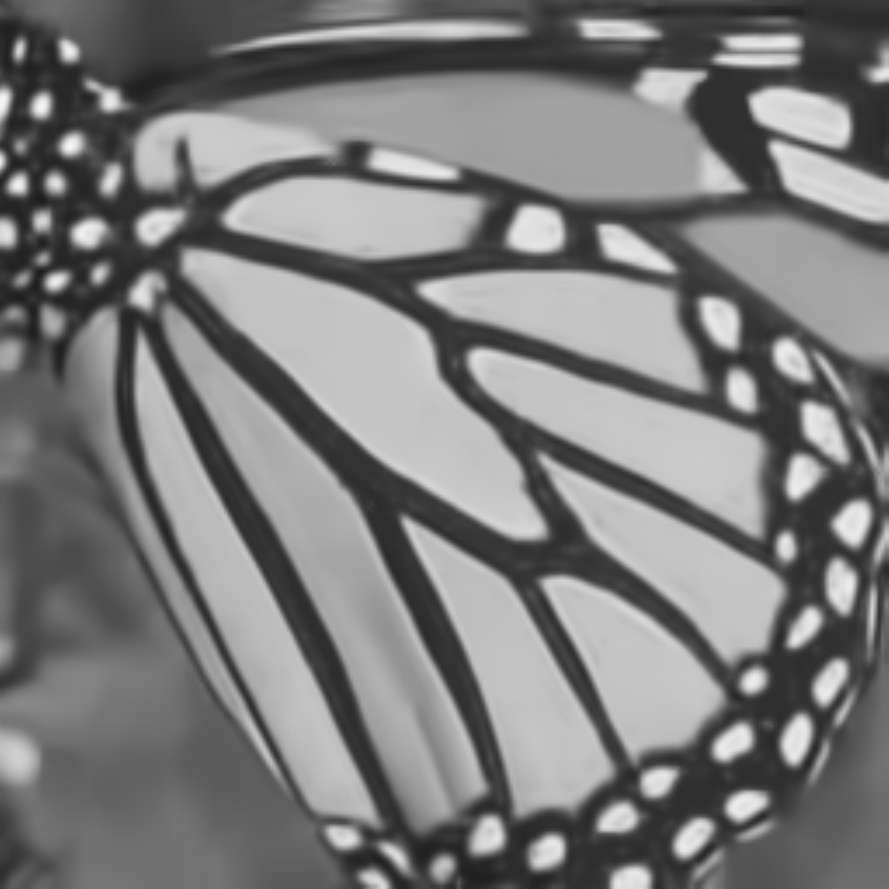}}
      \label{7c}\hspace{3mm}%
  \subfloat[Proposed (non-blind)]{%
       \includegraphics[width=0.36\linewidth]{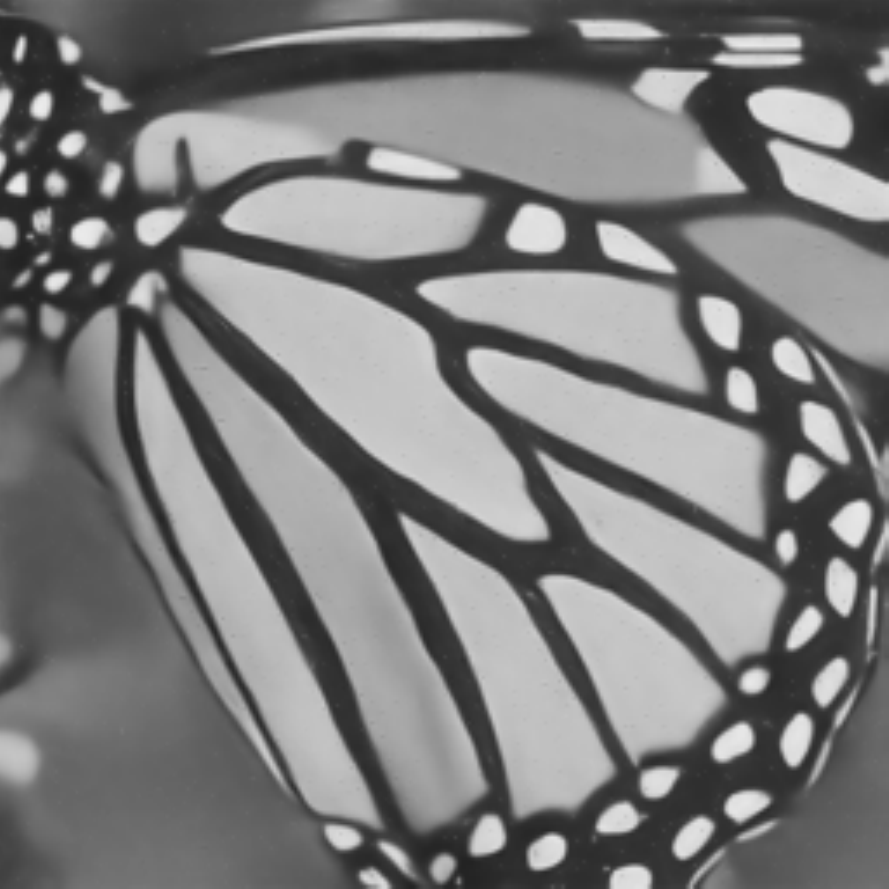}}
    \label{7d}
  \end{center}
  \caption{Result for untrained degradation: (a) the input is noised by salt-and-pepper with the density of 0.05, which is not trained. (b) DnCNN-3 failed to reduce the noise. (c) BM3D successfully reduced the noise. (d) our model reduces the noise with degradation attribute of Gaussian noise $\sigma=50/255$. }
  \label{fig:saltnpepper}
\end{figure}






\begin{table}[t!]
\begin{center}

\caption{Restoration results for salt-and-pepper noise}

\label{tab:psnr_snp}

\begin{tabular}{|c|c|c|c|} \hline
 & DnCNN-3 & BM3D & Proposed \\ \hline
Dataset & PSNR / SSIM & PSNR / SSIM & PSNR / SSIM \\ \hline
Set5 & 21.41 / 0.5319 & {\bf 27.87 / 0.7943} & 26.54 / 0.7301 \\ \hline
Set14 & 22.07 / 0.5652 & {\bf 26.50 / 0.7079} & 25.22 / 0.6320 \\ \hline
LIVE1 & 22.05 / 0.5553 & {\bf 25.83 / 0.6937} & 25.00 / 0.6349 \\ \hline
\end{tabular}

\end{center}
\end{table}

\subsubsection{Upscaling after JPEG compression}

The images are firstly JPEG compressed with 10\% of quality, then the upscaling with 1\% is applied as the perturbation. The restoration performances are shown in Table \ref{tab:psnr_jpeg_upscaling}.
The proposed model surpasses DnCNN-3 for the degradation with perturbation, while they have the same level of deblocking performance for the degradation without perturbation as demonstrated in Table \ref{tab:psnr}.

\begin{table}[t!]
\begin{center}

\caption{Restoration results for JPEG + upscaling}

\label{tab:psnr_jpeg_upscaling}

\begin{tabular}{|c|c|c|c|} \hline
 & DnCNN-3 & BM3D & Proposed \\ \hline
Dataset & PSNR / SSIM & PSNR / SSIM & PSNR / SSIM \\ \hline
Set5 & 31.53 / 0.8797 & 29.56 / 0.8456 & {\bf 31.64 / 0.8817} \\ \hline
Set14 & 29.87 / 0.8392 & 27.84 / 0.7698 & {\bf 29.97 / 0.8410} \\ \hline
LIVE1 & 29.20 / 0.8324 & 27.05 / 0.7488 & {\bf 29.28 / 0.8334} \\ \hline
\end{tabular}

\end{center}
\end{table}

\subsection{Controlling degree of restoration}

An interesting property of the proposed model is that restoration strength can be adjusted through the degradation attribute channels. Furthermore, the strength can be designated pixel-by-pixel because the attribute channels have exact same size as the input image.

\begin{figure}[t]
  \captionsetup{farskip=0pt}
    \begin{center}
  \subfloat[Original]{%
       \includegraphics[width=0.36\linewidth]{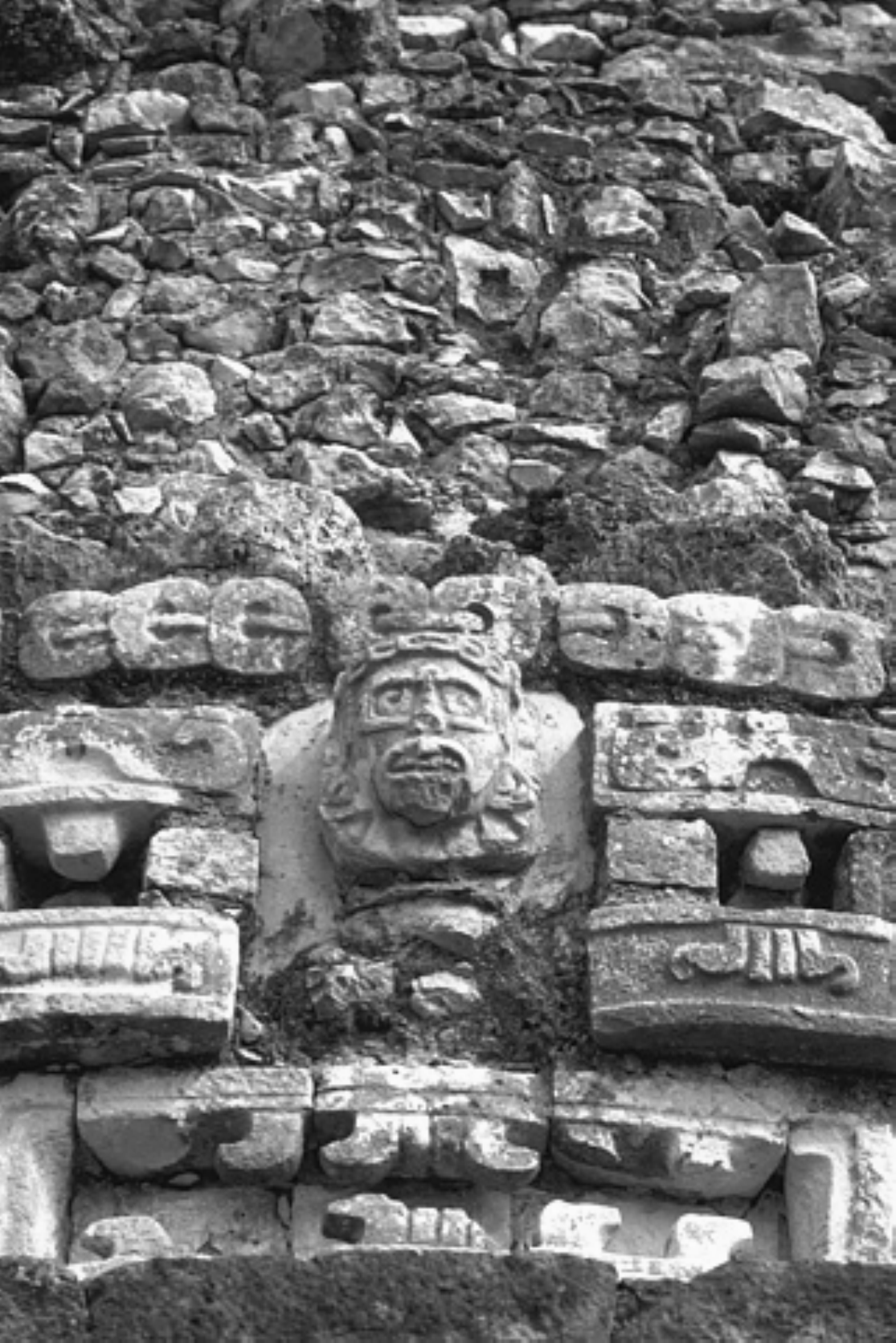}}
    \label{6a}\hspace{4mm}%
  \subfloat[Noisy]{%
       \includegraphics[width=0.36\linewidth]{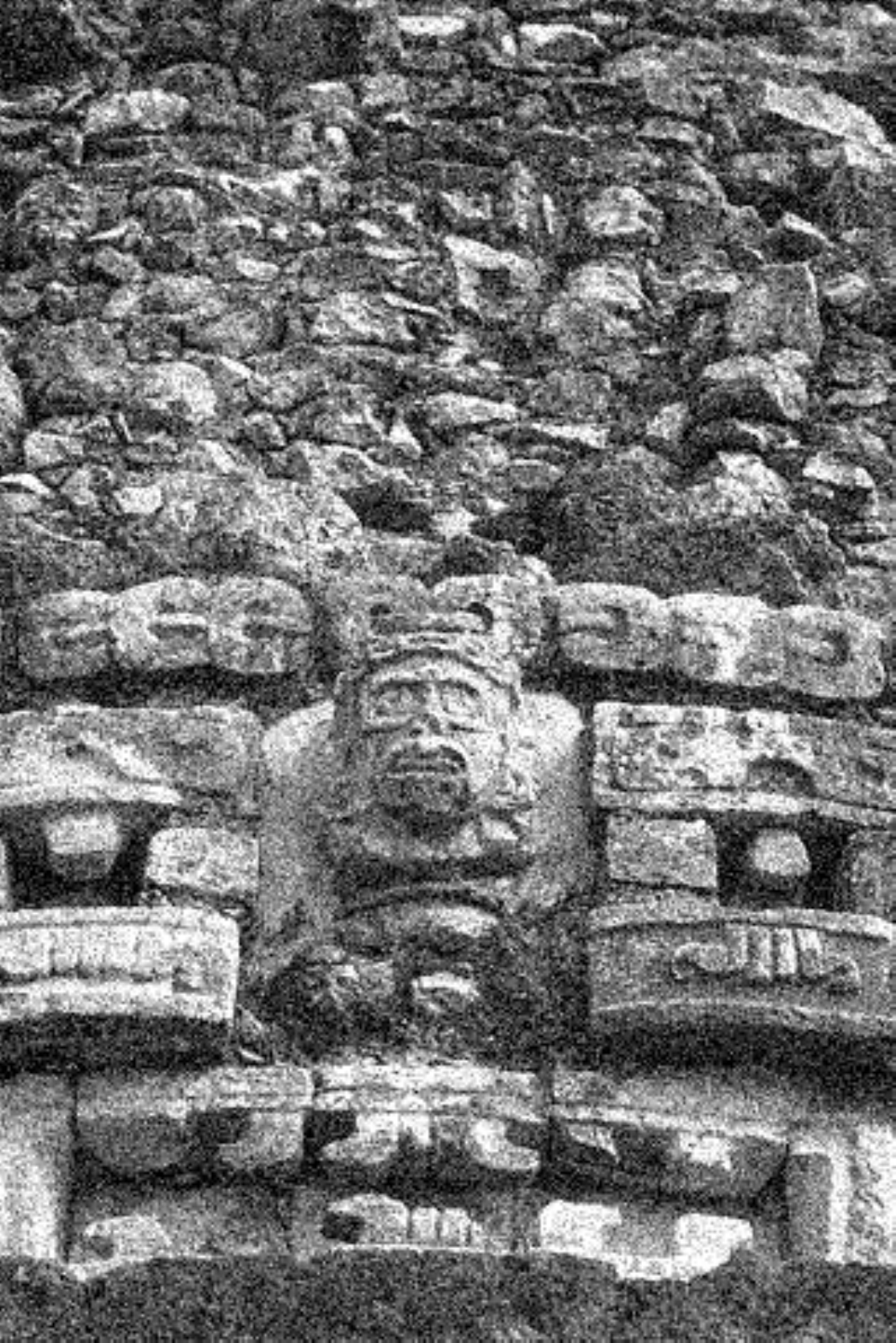}}
    \label{6b}
    \\
  \subfloat[Degradation attribute]{%
       \includegraphics[width=0.36\linewidth]{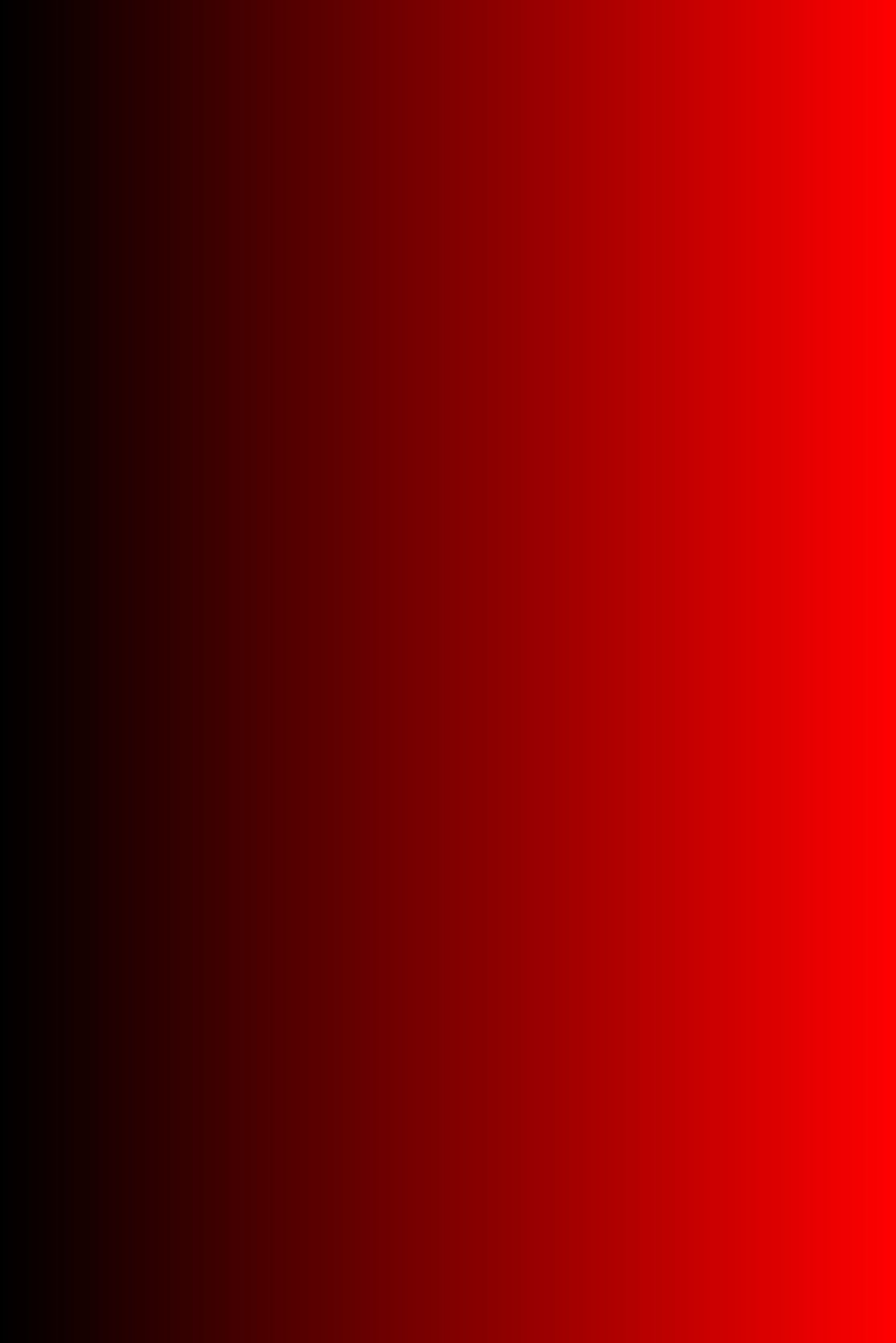}}
    \label{6c}\hspace{4mm}%
  \subfloat[Restored]{%
       \includegraphics[width=0.36\linewidth]{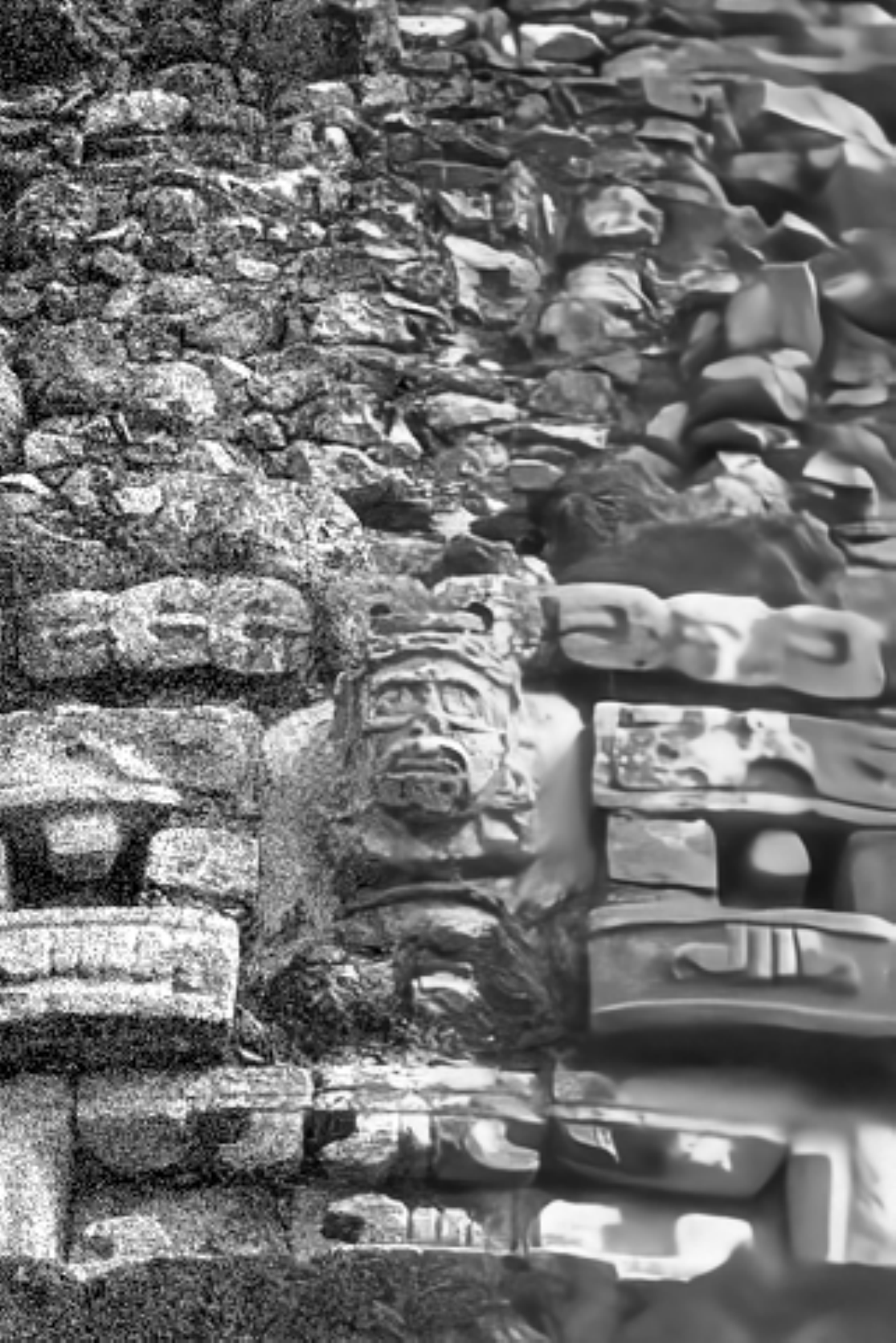}}
    \label{6d}
  \end{center}
  \caption{Non-blind restoration by region: (a) is the original image. (b) is Gaussian noised with $\sigma = 30/255$. (c) shows degradation attribute channels where the red component represents values for Gaussian noise level. (d) is the non-blind restoration image of (b); noise remains on the left side and details disappear on the right side. }
  \label{fig:drift_gd}
\end{figure}

Figure \ref{fig:drift_gd} shows denoised results for different values in degradation channels by region.
The input is noised with $\sigma=30/255$. In the degradation attribute channels, values for Gaussian noise level are gradual in horizonal direction. As a result, the restored image appears still noisy on the left side and smooth on the right side. As the values in the degradation attribute increases, the noise and high frequency detail disappears.
We consider that this property is particularly useful for human interactive image restoration.

\section{Conclusion}

In this paper, we proposed a non-blind CNN-based image restoration processor. By adding attribute channels indicating degradation property to the input channel, non-blind image restoration is realized.
Through experiments, we confirmed that the proposed model can be controlled by the input attribute channels, and restoration from untrained degradation model is achieved.

\bibliographystyle{IEEEtranKonno}
\bibliography{IEEEabrv,reference}

\end{document}